\documentclass{article}
\usepackage{ijcai19}

\usepackage{times}
\usepackage{soul}
\usepackage{url}
\usepackage[hidelinks]{hyperref}
\usepackage[utf8]{inputenc}
\usepackage[small]{caption}
\usepackage{graphicx}
\usepackage{amsmath}
\usepackage{booktabs}
\usepackage{algorithm}
\usepackage{algorithmic}
\title{Pixelation is NOT Done in Videos Yet}
\author{
Jizhe Zhou$^1$\footnote{Contact Author}\and
ChiMan Pun$^1$\And
Yingyu Wang$^1$\\
\affiliations
$^1$Dept. of Computer Information Science, FST, University of Macao\\
Taipa, Macao SAR\\
\emails
\{yb87409,cmpun\}@um.edu.mo,
}

\begin{document}

\maketitle

\begin{abstract}
This paper introduces an algorithm to protect the privacy of individuals in streaming video data by blurring faces such that face cannot be reliably recognized. This thwarts any possible face recognition, but because all facial details are obscured, the result is of limited use. We propose a new clustering algorithm to create raw trajectories for detected faces. Associating faces across frames to form trajectories, it auto-generates cluster number and discovers new clusters through deep feature and position aggregated affinities. We introduce a a Gaussian Process to refine the raw trajectories. We conducted an online experiment with 47 participants to evaluate the effectiveness of face blurring compared to the original photo (as-is), and users’ experience (satisfaction, information sufficiency, enjoyment, social presence, and filter likeability).

%%%%%%%%%%%%%%%old version submitted to IJCAI2019%%%%%%%%%%%%%%%%%%%%%%
%With the prevailing of online video streaming (especially outdoor-streaming), we develop a new tool called face pixelation in live-streaming (FPLV) to generate automatic filtering of personal privacy. Initiating from the frame level in which pixelation happens, FPLV presents a frame-to-video method for fast and accurately identifying and blurring an unknown number of faces during unconstrained live-streaming. Employing image-based face detection and recognition networks on every frame, we propose a new positioned incremental affinity propagation (PIAP) clustering algorithm to create raw trajectories for detected faces among consecutive frames. Associating faces across frames to form trajectories, PIAP auto-generates cluster number and discovers new clusters through deep feature and position aggregated affinities. Due to the variations of casting qualities, illumination, head pose, and partial occlusion, the created raw trajectories might be intermittent and unreliable. To address this challenge, we introduce a proposal net for loosed face detection with a Gaussian kernel empirical likelihood to compensate the deep feature insufficiency and recursively refine the raw trajectories. A Gaussian filter is laid on refined trajectories for final pixelation. FPLV obtains satisfying accuracy and real-time performances under streaming video data we collected.
%%%%%%%%%%%%%%%%%%%%%%%%%%%%%%%%%%%%%%%%%%%%%%%%%%%%%%%%%%%%%%%%%%%%%%%%%%

\end{abstract}

\section{Introduction}

Two experiments are reported that assess how well the identity of highly familiar (famous) faces can be masked from short naturalistic television clips. Recognition of identity was made more difficult by either pixelating  or blurringthe viewed face. Participants were asked to identify faces from both moving and static clips. Results indicated that participants were still able to recognize some of the viewed faces, despite these image degradations. In addition, moving images of faces were recognized better than static ones. 
\par
The interest resulting from these complaints has led to much work being carried out on what appeared to be a potentially large problem. This work has precipitated recommendations and specifications from many quarters which in many cases are overcritical and rigid and do not take into account the great flexibility of the visual system and the need for mobility to maintain postural comfort. In this paper the VDT issue is considered with reference to: the known physiology of the visual and postural mechanisms; the psychological factors such as fatigue, boredom, stress and performance of operators; and the clinical aspects of ocular comfort.
\par
Member States shall protect the
fundamental rights and freedoms of natural
persons, and in particular their right to privacy,
with respect to the processing of personal
data.
\par
To explore the potential blurring has as a privacy-enhancing technology for OSN photos, we conducted an online experiment with 47 participants to evaluate the effectiveness of face blurring compared to the original photo (as-is), and users’ experience (satisfaction, information sufficiency, enjoyment, social presence, and filter likeability). Users’ experience ratings for face blurring were positive, indicating blurring may be an acceptable way to modify photos from the users’ perspective. However, from a privacy-enhancement perspective, while face blurring may be useful in some situations, such as those where the person in the photo is unknown to the viewer, in other cases, such as in an OSN where the person in the image is known to the viewer, face blurring does not provide privacy protection.
\par

\iffalse   %这里分slice 好像不用了 说明n buffer就行  放倒后边说
By asking a $n$ frames long buffering time at the very beginning of a live-stream, we can cut the entire streaming into video segments or slices with each of them lasting $n$ frames long. Apparently, the longer segments offer more information for accuracy but also reduce more of the efficiency. In this paper, FPLV cuts every 90 frames into a segment and leverages the whole inner-segment context to reach accurate and fast face pixelation.
\fi

\section{Related Work}

Naturally, one may think pixelation could be roughly realized just by fine-tuning deep face detection and recognition CNNs alternately for live-stream. Such a view can be described as finding the answers to "where are the people in this frame?" and "which of them is the streamer?" frame by frame. However, put the accuracy aside, deep CNN based face recognition network builds a classification algorithm which demands a large number of streamers' faces as a prior to work properly.

\par 

Current solutions of MOT or MFT in unconstrained videos assumes some kind of priors including numbers of people shows up in videos, initial positions of the people, no fast motions, no camera movements, and high recording resolution to work correctly. Latest studies on MFT $R$ manage to exclude as many priors as possible but require the full-length video in advance for analysis $R$.

Realized through continuously motion prediction and corrections, object tracking including tracking by detection algorithm heavily suffers from these characteristics of streaming videos and struggles to handle the tracklet linking and drifting problems.

Notably, blinking mosaics are extremely unacceptable as they primarily lead to privacy leakage and will instantly attract viewers' attention. Then they quickly turn to be very annoying. Besides, the recognition model trained on image datasets cannot represent the temporal connection of the same people across adjacent frames. As a result, it aggravated the blinking phenomenon.

Therefore, after the face detection and recognition stage, we aim to build a novel clustering algorithm that provides pair-wise similarities for faces regarding both deep feature and motion information of faces. With an unknown number of faces, affinity propagation (AP) clustering becomes the most suitable choice. However, like other clustering algorithms, AP is designed for discovering patterns in static data. Due to the high-speed process requirements of video live-streaming, we extended the AP clustering into an incremental way to dynamically process the incoming frames and link the same faces through frames to generate the raw face trajectories. Then, rather than a single face detector, another face proposal net is involved instantly as a supplement. More precisely, affected by the video resolution and motion blur, ~\cite{zhang2016joint} is applied with a more strict threshold to reject the false positives for higher precision. In order to reach a high rate of accuracy, we manage another shallow CNN to propose suspicious face areas at a top re-call rate to compensate the detection lost. Combining a GP model with raw trajectories, we deny the unexpected ones and receive the correct ones for refinement. The final face pixelation is realized by overlapping Gaussian Filters on target refined trajectories.

\section{Proposed Method}

Considering all the pros and cons need to be handled, we propose a   framework, which contains three core stages:$\textbf{(1) Face Detection, Alignment, and Recognition.}$ As discussed previously, live-streaming videos are made up of frames. Intuitively, referring to some well-established face detection and recognition algorithms, we can automate the pixelation action within any individual frames. Therefore, in this paper, we start with $MTCNN$. This face detector follows the way of applying multi-layer, fast and shallow CNN in a cascading structure. Detected faces are cropped and aligned to frontal pose through affine transformation. Instead of directly using a recognition algorithm, we build a triple-loss styled deep face embedding CNN according to ~\cite{schroff2015facenet}. Initially designed in face recognition area, ~\cite{schroff2015facenet} extracts high-level features for each face and offers us a reliable basis in distinguishing faces. $\textbf{(2) Face Trajectories Forming.}$ Since each face is independently identified in (1), we shall recursively link same faces across frames in a sequential order plot the raw trajectory for a particular face. Since the number of people shows up in streaming cannot be known in advance, and we need a fast enough algorithm for real-time consideration, affinity propagation clustering algorithm is chosen for linkage. More specifically, an incremental affinity propagation clustering algorithm associated with the position is developed to organize such linkage across frames in real-time and eliminate false positives occurred in the detection stage.$\textbf {(3) Trajectory Refinement.}$ Despite the eliminated false positives, the detection caused false negatives are the real challenge we incurred. Normally, image-based CNN is very sensitive to resolutions and motion blur since they are designed and trained on high-quality images. Thus, we will always get intermittent trajectories for every faces under live-streaming conditions. To solve this issue, a Gaussian Process model is designed to capture the extra richness of video data through context and compensate the deep learning offered features' insufficiency. Our refinement model vanishes the discontinuity and prevents personal privacy from leaking.

\subsection{Detection and Recognition within Video Segments}
We start by deploying the face detection and recognition network. In this paper, MTCNN proposed by~\cite{zhang2016joint}, and Facenet proposed by~\cite{schroff2015facenet} process every frame in turn. Note that the detection and recognition network could be substituted by other recent published state-of-the-art work like PyrimaidBox~\cite{tang2018pyramidbox} \& Deepface~\cite{taigman2014deepface}.
%%%对于如何elaborately 裁剪face的描述

\subsection{Raw Face Trajectories Forming}
%%%put the picture I always draw here... several faces in %%%different frames. 10 faces in 4 frames belong to 4 people.....

To accomplish the face pixelation in videos based on what we have done in the above stage, a simple and direct way could be adapting a K-NN like classification algorithm in every detached frame to identify the streamers' faces and blur the rest ones. Clearly, such method comes with troublesome drawbacks: $\textbf{How Close Two Faces Need to be Recognized as a }$
$\textbf{Same Person}$. Whatever classification algorithms are applied, they concentrate on finding a proper threshold in distinguish two faces' vector from each other. Deep CNN based algorithms are very sensitive to datasets and environment conditions, and K-NN styled algorithms require as many pre-defined faces for references as possible, their classification mechanism doesn't hold under unconstrained videos. In contract, clustering algorithms offer us the way to faces are belong to a same person.

\iffalse
\begin{figure}[ht]
\centering
\includegraphics[scale=0.45]{1.png}
\caption{Our PIAP can discovery new cluster automatically. In this toy example, new faces arrive in (c), they all come from a new cluster. Assignments is implemented in (d), and message-passing continues in (e)∼(h). PIAP reconverges in (i), and the final result of new face clusters is shown in (j).}
\label{fig:label}
\end{figure}
\fi

\begin{algorithm}[tb]
\caption{Positioned Incremental Affinity Propagation}
\label{alg:algorithm}
\textbf{Input}: $R_{t-t'}$,$A_{t-t'}$,$\hat{c_{t-t'}}, Z$\\
\textbf{Output}: $R_{t},A_{t},\hat{c_{t}}$
\begin{algorithmic}[1] %[1] enables line numbers
\WHILE {not end of a live-streaming}
\IF {the first video segment of a live-stream}
\STATE Assign zeros to all responsibilities and availabilities.
%\ELSE
%\STATE Compute responsibilities and availabilities for $Z$ according to equation (5) and (7).
%\STATE Extend responsibilities matrix $R_{t-t'}$ to $R_{t}$, and availabilities $A_{t-t'}$ to $A_{t}$.
\ENDIF
%\STATE Message-passing according to equation (2), (3) and (8) until convergence.
%\STATE Compute $\hat{c_t}$ as equation (4).
\ENDWHILE
\end{algorithmic}
\end{algorithm}

\subsection{Trajectory Refinement}
For the sake of simplicity, mean value is all set to zero in our work and the covariance equals $\sigma_{z^{'}}^2 k(z'_p,z'_q;\theta)$. $f(z,\theta) \sim \mathcal{N}(0, \Sigma)$ where $\Sigma$ is the covariance matrix with $pq^{th}$ element equal to $k(z'_p,z'_q;\theta)$. As our attention is focused on the the hyparameter $\theta$ for dimension reduction, and we want to compute the Gaussian likelihood function $L_{z^{'}}$ respect to $\theta$ as:
%Following the widely used kernel and since $z'$ and $z$ are trained with deep CNN, $\theta$ is defined by correlation matrix with the radius basis function (RBF) kernel. 
%
%\begin{equation}
%   k_f(z_p,z_q)=\theta_{rbf} \exp (-\frac{\theta_{band}}{2}(z_p-z_q)^{T}(z_p-z_{q}))
%\end{equation}
\begin{equation}
    L_{z'}=\ln{p(\theta|z,z')}=-\frac{dn}{2}\ln{2\pi}-\frac{d}{2} \det{\Sigma} -\frac{1}{2}tr(z'\Sigma^{-1}(z')^{T})
\end{equation}

\par

%%%%%%%%%%%%%%%%对于图片描述
%\par
%As demonstrated in $Figure X$, each frame is processed by face detection and recognition algorithms independently, we can %write:
%\begin{equation}
%   p(x|z_1,z_2,...,z_n)= \prod_{i=1}^{d}p(x_i|z_1,z_2,...,z_n)
%\end{equation}
%%%%%%%%%%%%%%%%%

%%%%%%%%%%%%%
%%%%%%%%%%%%
%%%%%%%%%%%%
%%%%%%%%%%%%算法改动

The entry of $K$ is given like:
\begin{equation}
   K_{p,q}=k_{f}(z_p,z_q)+\sigma_x^{2}I
\end{equation}
%%$\sigma_x^{2}\delta_{{p}{q}}$ is the noise when obtaining data $z'$. \\
Clearly, the marginal likelihood for each dimension of $z'$ is:

\begin{equation}
    \begin{aligned}
   p(x_{:,d}|z, \theta) &= \int p(x_{:,d}|z,f)p(f)df \\ &=N(x_{:,d}|0,K(z,\theta))
   \end{aligned}
\end{equation}
Easily, as every frame is process by detection and recognition algorithm independently, i.i.d assumption stands, and the likelihood for observed $z'$ is:
\begin{equation}
    \begin{aligned}
    p(x|z,\theta) &=\prod_{i=1}^{d}p(x_{:,d}|z,\theta) \\ &= \prod_{i=1}^{d}\frac{1}{2\pi^{\frac{n}{2}}{|K|}^\frac{1}{2}}\exp{(-\frac{1}{2}x_{:,d}^{T}K^{-1}x_{:,d})}
   \end{aligned}
\end{equation}
(14) is a product of $d$ independent Gaussian processes. Omit writing hyparameter $\theta$ in the following computation for simplicity, we can reformulate the likelihood to omit the $\theta$ by considering:
\begin{equation}
   p(x|z)=\frac{1}{2\pi^{\frac{n}{2}}{|K|}^\frac{1}{2}}\exp{(-\frac{1}{2}tr(K^{-1}XX^{T}))}
\end{equation}
Here $p(x|z)$ is the likelihood.
As $z$ and $z'$ are current observations, according to Bayesian rule, $p(z)$ and $p(x)$ can be excluded from computation. And maximizing the posterior $p(z|x)$ is equal to maximizing the log likelihood of $p(x|z)$ straightforwardly:
\begin{equation}
   \{\hat{z},\hat{\theta}\}=\arg\max_{\theta} \{\ln{p(z|x)} \}
\end{equation}

Or equivalently
\begin{equation}
   \{\hat{z},\hat{\theta}\}=\arg\max_{x,\theta} \{\ln{p(x|z)+\ln{p(z)}} \}
\end{equation}

Since each frame is processed by face detection and recognition algorithms independently, we can write:
\begin{equation}
   p(x|z_1,z_2,...,z_n)= \prod_{i=1}^{d}p(x_i|z_1,z_2,...,z_n)
\end{equation}
So we have:
\begin{equation}
   \ln{p(x|z)}= \sum_{i=1}^{d}\ln{p(x_:,i|z)}
\end{equation}
(16) is the log likelihood function of $z'$. $\ln{p(x|z)}$ as a product of $d$ independent Gaussian processes, and each process is related to a different dimension of the set of observations $z$. According to (15), we can rewrite (19) to:
\begin{equation}
   L_x=\ln{p(x|z)}=-\frac{dn}{2}\ln{2\pi}-\frac{d}{2}|K|-\frac{1}{2}tr(K^{-1}xx^{T})
\end{equation}
From (17), as $z$ and $z'$ are the observation we have, instead of complicated E-M algorithm, we instantly apply MLE on (20), and we can get the value of hyperparameter $\theta$:
\begin{equation}
 \frac{\partial{L_x}}{\partial {\theta}}=\frac{\partial{L_x}}{\partial {K}}\frac{\partial{K}}{\partial {\theta}}=0
\end{equation}
Then, the gradients of the kernel matrices with respect to the hyperparameters can be computed by:
\begin{equation}
    \left\{  
             \begin{array}{lr}  
             \frac{\partial{K}}{\partial {\theta_{rbf}}}=\frac{1}{\theta_{rbf}}K\\
             \\
             \frac{\partial{K}}{\partial {\theta_{band}}}=K \hat{K} \\
             \\
             \frac{\partial{K}}{\partial {\sigma_x^2}}=I  
             \end{array}  
\right.  
\end{equation}
Where matrix $\hat{K}=-\frac{1}{2}(z_p-z_q)^{T}(z_p-z_q)$. And the hyperparameter $\theta$ is the result of characteristic after GP models for dimension reduction.

%%%%%%%%%%%%%%%%
%%%%%%%%%%%%%%%%
%%%%%%%%%%%%%%%%
%%%%%%%%%%%%%%%%
%%%%%%%%%%%%%%%%重新调整的ELR算法  将 GP超级简化
Following the mechanism of $GP-LVM$, we easily deduce the jointly likelihood for (10): 
\begin{equation}
    \begin{aligned}
   p(x_{:,d}|z, \theta) &= \int p(x_{:,d}|z,f)p(f)df \\ &=N(x_{:,d}|0,K(z,\theta))
   \end{aligned}
\end{equation}
As both $z$ and $z'$ are observations, instead of E-M algorithm in original $GP-LVM$, the maximum likelihood estimates (MLE) of $\theta$ can be obtained by maximizing the logarithm of the joint likelihood with Bayesian rule:
\begin{equation}
   \hat{\theta}=\arg\max_{\theta} \{\ln{p(z|x, \theta)} \}
\end{equation}
and
\begin{equation}
   p(x|z)=\frac{1}{2\pi^{\frac{n}{2}}{|K|}^\frac{1}{2}}\exp{(-\frac{1}{2}tr(K^{-1}XX^{T}))}
\end{equation}
\par
This GP model may sufficient enough for false positives rejection under most cases $R$. However, in live-streaming, we process on short video segments and the small amount of frames with high dimensional feature vectors may not supports such a GP model functions well under Central Limit Rule. Therefore, a parametric solution of empirical likelihood ratio through Wilk's Theorem is used to solve such a prior-less two-sample test.  
\iffalse
\subsection{Length of Papers}

All paper {\em submissions} must have a maximum of six pages, plus at most one for references. The seventh page cannot contain {\bf anything} other than references.

The length rules may change for final camera-ready versions of accepted papers, and will differ between tracks. Some tracks may include only references in the last page, whereas others allow for any content in all pages. Similarly, some tracks allow you to buy a few extra pages should you want to, whereas others don't.

If your paper is accepted, please carefully read the notifications you receive, and check the proceedings submission information website\footnote{\url{https://proceedings.ijcai.org/info}} to know how many pages you can finally use (and whether there is a special references-only page or not). That website holds the most up-to-date information regarding paper length limits at all times.

\subsection{Word Processing Software}

As detailed below, IJCAI has prepared and made available a set of
\LaTeX{} macros and a Microsoft Word template for use in formatting
your paper. If you are using some other word processing software, please follow the format instructions given below and ensure that your final paper looks as much like this sample as possible.
\fi
\
\section{Experiments}
\subsection{Details}
\paragraph{Dataset:}As face pixelation problem in live video streaming is not studied before, there are no available dataset and benchmark tests for reference. We collected and built a live-streaming video dataset from YouTube and Facebook platforms and manually pixelated the faces for comparison. Limited by the current source, we select four short and relatively not complicated live-streaming fragments from our dataset for labeling. These fragments contain four records of live-stream videos total lasting 1012 seconds. Since they all are streamed in 30FPS, each fragment contains 7500+ frames on average. As demonstrated in Table 1, these fragments are further divided into two groups according to resolution and number of people showed up in a stream. Live-stream videos with at least 720p resolution are marked as high-resolution $H$, and the rest are low-resolution $L$. Similarly, live-stream contains more than two people is sophisticated $S$ and the rest are naive $N$.
\begin{table}
\centering
\begin{tabular}{cccc}
\toprule 
Property  &Frames & Resolution & Number of people \\
\midrule
$HS$      &6753  & 720p &4  \\
$LS$      &6604 & 480p &3  \\
$HN$      &7721 & 1080p &2 \\
$LN$      &9401 & 480p & 1 \\
\bottomrule
\end{tabular}
\caption{Details of the live-stream dataset}
%\label{tab:plain}
\end{table}

\paragraph{Evaluation metrics:}
\iffalse
(1) Clustering. Clustering purity is the only off the shelf metric we could use. Comparing to $R$, we employ weighted clustering purity (WCP) to evaluates the performance of PIAP. The clusters' number is also added as a comparison to check the performance of PIAP. (2)
\fi
Set our manual pixelation as the baseline, we adopt the algorithms including MFT and state-of-the-art face detection and recognition to face pixelation tasks and compute their precision and recall rate of face mosaics. From Table 2 and Table 3, the state-of-the-art detection and recognition outputs the highest precision in $HN$ but with a meager recall rate which means it misses lots of detections and the mosaics fast blink on faces. HOG, VGG-face19, and Tacking-Learning-Detection (TLD) algorithms are adopted to pixelation task for comparison.

\iffalse
\begin{table}
\centering
\begin{tabular}{ccccc}
\toprule 
Method  &$HS$ & $LS$ & $HN$ &$LN$ \\
\hline
HOG     &0.31  &0.30 &0.32 & 0.26  \\
VGG-face      &0.34 &0.33 &0.43 & 0.30  \\
Facenet &0.68 &0.57 &0.82 & 0.69  \\
Deepface &0.74 &0.68 &0.86 & 0.77  \\
\midrule
PIAP      & \textbf{0.77} &\textbf{0.73}  &\textbf{0.89} & \textbf{0.88}\\
Cluster number &4/4&3/3&2/2&1/1\\
\bottomrule
\end{tabular}
\caption{WCP comparisons on live-stream video dataset}
%\label{tab:plain}
\end{table}
\fi
\begin{table}
\centering
\begin{tabular}{ccccc}
\toprule 
Method  &$HS$ & $LS$ & $HN$ &$LN$ \\
\hline
YouTube     &0.85   &0.78 &0.84  &0.89\\
Azure      &0.77  &0.53 &0.77  &0.75\\
TLD        &0.79 &0.72 &0.76 & 0.72 \\
PyrimidBox+Deepface &0.92 &0.84 &\textbf{0.98} & 0.82 \\
\midrule
FPLV       & \textbf{0.93} &\textbf{0.91}  &0.95 & \textbf{0.90}\\
\bottomrule
\end{tabular}
\caption{Precision on live-stream video dataset}
%\label{tab:plain}
\end{table}
\begin{table}
\centering
\begin{tabular}{ccccc}
\toprule 
Method  &$HS$ & $LS$ & $HN$ &$LN$ \\
\hline
YouTube     &0.77  &0.83 &0.88 &\textbf{0.91} \\
Azure      &0.73  &0.67 &0.84  &0.88\\
TLD        &0.75 &0.72 &0.78 & 0.81 \\
Pyrimid+Deepface &0.61 &0.44 &0.66 & 0.47 \\
\midrule
FPLV       & \textbf{0.77}  &\textbf{0.81}  &\textbf{0.87} &\textbf{0.88} \\
\bottomrule
\end{tabular}
\caption{Recall rate on live-stream video dataset}
%\label{tab:plain}
\end{table}
\subsection{Results on live video streaming data}
Self-evidently, FPLV generates best face pixelation results listed in above tables. Note that although YouTube studio performs close to our FPLV under most test, the face pixelation tool of YouTube and Azure cannot process the video in real-time. Figure 4 demonstrates a partial result\footnote{More video results on:\href{https://knightzjz.github.io/}{fplv.github.io}} of FPLV comparing to YouTube Studio. In a live-stream of Jam Hsiao on Facebook, the upper line results of FPLV well handled the pixelation task when two faces overlap each other. The bottom line generates by YouTube through offline face tracking algorithms fails such a pixelation as other person's mosaics are incorrectly overlaid on streamer's face.
\subsection{Efficiency}
The main cost of FPLV is on the face embedding algorithm. The embedding for one face takes 10-15ms on our i7-7800X, GTX1080, 32G RAM machine, and the face detection including compensation for a frame take 3ms. Another time-costing part is the initialization of IPAP; the initialization process takes 10-30ms depending on the case. And each incremental propagation loop takes 3-5ms. So FPLV can satisfy the real-time efficiency requirement, and under extreme circumstances that contain a lot of faces, we can trick the detection and recognition to jump among frames for efficiency.
\begin{figure}[ht]
\centering
\includegraphics[scale=0.30]{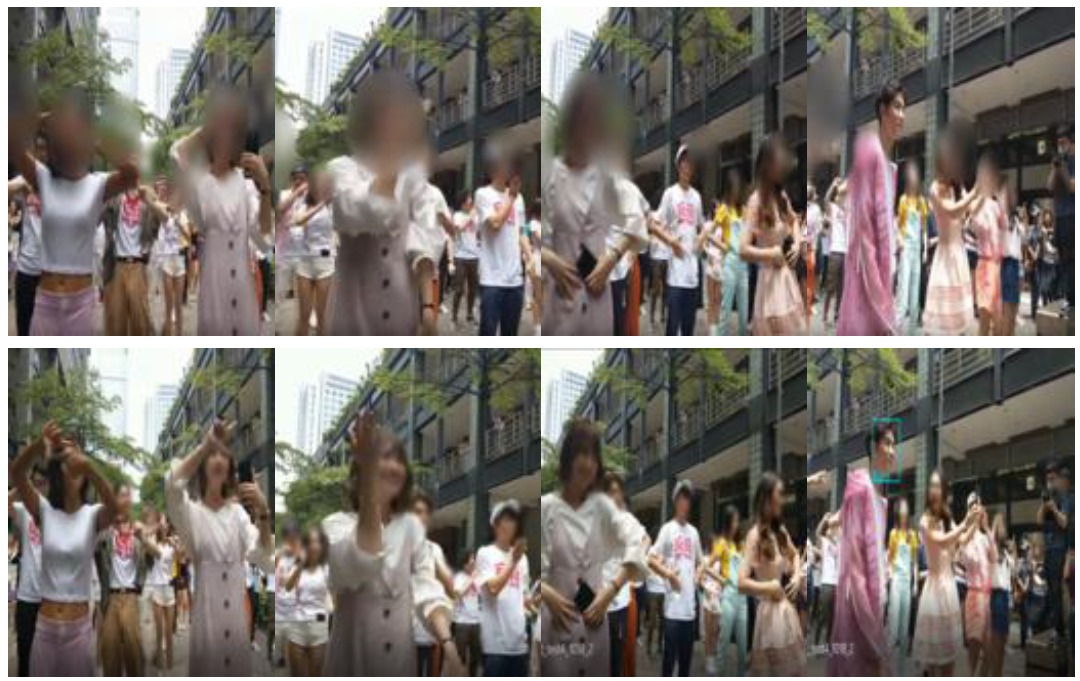}
\caption{Results comparison between FPLV and YouTube Studio}
\label{fig:label}
\end{figure}
\par 
\section{Conclusions}
According to our knowledge, we are the first to address the face pixelation problem in live video streaming by building the proposed FPLV. FPLV is already surpassing offline tools offered by YouTube and Microsoft and becomes applicable in real life scenarios. FPLV can achieve high accuracy and real-time performances on the dataset we collected. We will extend FPLV to behave better on low-resolution data in the future.

\bibliographystyle{named}
\bibliography{ijcai19}

\end{document}